\documentclass[11pt,a4paper]{article}
\usepackage[hyperref]{acl2018}
\aclfinalcopy 
\def\aclpaperid{1348} 

\usepackage{times}
\usepackage{url}
\usepackage{latexsym}

\usepackage{listings}

\usepackage{booktabs}
\usepackage{tabularx,multirow}
\usepackage[T1]{fontenc}
\usepackage[utf8]{inputenc}
\usepackage[english]{babel}
\usepackage{blindtext}
\usepackage{amsmath}
\usepackage{amsfonts}
\usepackage{graphicx}
\usepackage{natbib}
\usepackage{gb4e}  

\noautomath

\newcommand{\beq}{\begin{equation}}
\newcommand{\eeq}{\end{equation}}

\definecolor{apgreen}{RGB}{51, 139, 0}

\usepackage{color}
\definecolor{NiceBlue}{RGB}{11, 102, 163}
\definecolor{SlightRed}{RGB}{249,38,114}
\usepackage{textcomp} 
\lstdefinelanguage{DemoExample}
{ basicstyle=\footnotesize\ttfamily,
  commentstyle=\color{SlightRed}\rmfamily\itshape,
  stringstyle=\color{NiceBlue},
  morecomment=[s]{/*}{*/},
  morestring=[b]'
}

\begin{document}

\author{Andre Cianflone \thanks{\hspace{2mm}Authors (listed in alphabetical order) contributed equally.}\qquad Yulan Feng $^*$ \qquad Jad Kabbara $^*$ \qquad Jackie Chi Kit Cheung\\ \\
\hspace{-10mm}School of Computer Science \hspace{20mm} MILA\\
\hspace{12mm}McGill University \hspace{15mm}Montreal, QC, Canada\\ 
\hspace{-42mm}Montreal, QC, Canada\\ \\
{\tt \normalsize \{andre.cianflone@mail, \hspace{-2mm}yulan.feng@mail, \hspace{-2mm}jad@cs, \hspace{-2mm}jcheung@cs\}.mcgill.ca}
        }

\title{Let's do it ``again'': A First Computational Approach to Detecting Adverbial Presupposition Triggers}
\maketitle

\begin{abstract}
We introduce the task of predicting adverbial presupposition triggers such as \textit{also} and \textit{again}. Solving such a task requires detecting recurring or similar events in the discourse context, and has applications in natural language generation tasks such as summarization and dialogue systems. We create two new datasets for the task, derived from the Penn Treebank and the Annotated English Gigaword corpora, as well as a novel attention mechanism tailored to this task. Our attention mechanism augments a baseline recurrent neural network without the need for additional trainable parameters, minimizing the added computational cost of our mechanism. We demonstrate that our model statistically outperforms a number of baselines, including an LSTM-based language model.

\end{abstract}

\section{Introduction}
In pragmatics, presuppositions are assumptions or beliefs in the common ground between discourse participants when an utterance is made \cite{frege92,strawson-1950,stalnaker1973presuppositions,Stalnaker1998}, and are ubiquitous in naturally occurring discourses \cite{sep-presupposition}. Presuppositions underly spoken statements and written sentences and understanding them facilitates smooth communication. We refer to expressions that indicate the presence of presuppositions as \emph{presupposition triggers}. These include definite descriptions, factive verbs and certain adverbs, among others. For example, consider the following statements:
\begin{exe}
	\ex\label{ex1} John is going to the restaurant \textit{again}.
	\ex\label{ex2} John has been to the restaurant.
\end{exe}

(\ref{ex1}) is only appropriate in the context where (\ref{ex2}) is held to be true because of the presence of the presupposition trigger \textit{again}. One distinguishing characteristic of presupposition is that it is unaffected by negation of the presupposing context, unlike other semantic phenomena such as entailment and implicature. The negation of (\ref{ex1}), \textit{John is not going to the restaurant again.}, also presupposes (\ref{ex2}).

Our focus in this paper is on adverbial presupposition triggers such as \textit{again, also} and \textit{still}. Adverbial presupposition triggers indicate the recurrence, continuation, or termination of an event in the discourse context, or the presence of a similar event. In one study of presuppositional triggers in English journalistic texts \cite{khaleel2010analysis}, adverbial triggers were found to be the most commonly occurring presupposition triggers after existential triggers.\footnote{Presupposition of existence are triggered by possessive constructions, names or definite noun phrases.} 
Despite their frequency, there has been little work on these triggers in the computational literature from a statistical, corpus-driven perspective.

As a first step towards language technology systems capable of understanding and using presuppositions, we propose to investigate the detection of contexts in which these triggers can be used. This task constitutes an interesting testing ground for pragmatic reasoning, because the cues that are indicative of contexts containing recurring or similar events are complex and often span more than one sentence, as illustrated in Sentences~(\ref{ex1}) and~(\ref{ex2}). Moreover, such a task has immediate practical consequences. For example, in language generation applications such as summarization and dialogue systems, adding presuppositional triggers in contextually appropriate locations can improve the readability and coherence of the generated output.

We create two datasets based on the Penn Treebank corpus \cite{marcus-etal-1993} and the English Gigaword corpus \cite{gigaword}, extracting contexts that include presupposition triggers as well as other similar contexts that do not, in order to form a binary classification task. In creating our datasets, we consider a set of five target adverbs: \textit{too, again, also, still}, and \textit{yet}. We focus on these adverbs in our investigation because these triggers are well known in the existing linguistic literature and commonly triggering presuppositions. We control for a number of potential confounding factors, such as class balance, and the syntactic governor of the triggering adverb, so that models cannot exploit these correlating factors without any actual understanding of the presuppositional properties of the context. 

We test a number of standard baseline classifiers on these datasets, including a logistic regression model and deep learning methods based on recurrent neural networks (RNN) and convolutional neural networks (CNN).

In addition, we investigate the potential of attention-based deep learning models for detecting adverbial triggers. Attention is a promising approach to this task because it allows a model to weigh information from multiple points in the previous context and infer long-range dependencies in the data \cite{bahdanau2014neural}. For example, the model could learn to detect multiple instances involving \textit{John} and \textit{restaurants}, which would be a good indication that \textit{again} is appropriate in that context. Also, an attention-based RNN has achieved success in predicting article definiteness, which involves another class of presupposition triggers \cite{kabbara2016capturing}.

As another contribution, we introduce a new weighted pooling attention mechanism designed for predicting adverbial presupposition triggers. Our attention mechanism allows for a weighted averaging of our RNN hidden states where the weights are informed by the inputs, as opposed to a simple unweighted averaging. Our model uses a form  of self-attention \cite{paulus2017deep,vaswani2017attention}, where the input sequence acts as both the attention mechanism's query and key/value. Unlike other attention models, instead of simply averaging the scores to be weighted, our approach aggregates (learned) attention scores by learning a reweighting scheme of those scores through another level (dimension) of attention. Additionally, our mechanism does not introduce any new parameters when compared to our LSTM baseline, reducing its computational impact.

We compare our model using the novel attention mechanism against the baseline classifiers in terms of prediction accuracy. Our model outperforms these baselines for most of the triggers on the two datasets, achieving 82.42\% accuracy on predicting the adverb ``also'' on the Gigaword dataset.


The contributions of this work are as follows:
\begin{enumerate}
\item We introduce the task of predicting adverbial presupposition triggers.
\item We present new datasets for the task of detecting adverbial presupposition triggers, with a data extraction method that can be applied to other similar pre-processing tasks.
\item We develop a new attention mechanism in an RNN architecture that is appropriate for the prediction of adverbial presupposition triggers, and show that its use results in better prediction performance over a number of baselines without introducing additional parameters.
\end{enumerate}

\section{Related Work}
\subsection{Presupposition and pragmatic reasoning}

The discussion of presupposition can be traced back to Frege’s work on the philosophy of language \cite{frege92}, which later leads to the most commonly accepted view of presupposition called the Frege-Strawson theory \cite{Kaplan1970-KAPWIR,strawson-1950}. In this view, presuppositions are preconditions for sentences/statements to be true or false.
To the best of our knowledge, there is no previous computational work that directly investigates adverbial presupposition. However in the fields of semantics and pragmatics, there exist linguistic studies on presupposition that involve adverbs such as ``too'' and ``again'' (e.g., \cite{blutner2003optimality}, \cite{wangqiang}) as a pragmatic presupposition trigger.
Also relevant to our work is \cite{kabbara2016capturing}, which proposes using an attention-based LSTM network to predict noun phrase definiteness in English. Their work demonstrates the ability of these attention-based models to pick up on contextual cues for pragmatic reasoning. 


Many different classes of construction can trigger presupposition in an utterance, this includes but is not limited to stressed constituents, factive verbs, and implicative verbs \cite{zare_abbaspour_nia_2012}. In this work, we focus on the class of adverbial presupposition triggers.

Our task setup resembles the Cloze test used in psychology \cite{cloze-test,coleman,Rankin} and machine comprehension \cite{Riloff:2000:RQA:1117595.1117598}, which tests text comprehension via a fill-in-the-blanks task. We similarly pre-process our samples such that they are roughly the same length, and have equal numbers of negative samples as positive ones. However, we avoid replacing the deleted words with a blank, so that our model has no clue regarding the exact position of the possibly missing trigger. Another related work on the Children’s Book Test \cite{Hill2015TheGP} notes that memories that encode sub-sentential chunks (windows) of informative text seem to be most useful to neural networks when interpreting and modelling language. Their finding inspires us to run initial experiments with different context windows and tune the size of chunks according to the Logistic Regression results on the development set.

\subsection{Attention}
In the context of encoder-decoder models, attention weights are usually based on an energy measure of the previous decoder hidden state and encoder hidden states. Many variations on attention computation exist. Sukhbaatar et al.~\shortcite{sukhbaatar2015memnets} propose an attention mechanism conditioned on a query and applied to a document. To generate summaries, Paulus et al.~\shortcite{paulus2017deep} add an attention mechanism in the prediction layer, as opposed to the hidden states. Vaswani et al.~\shortcite{vaswani2017attention} suggest a model which learns an input representation by self-attending over inputs. While these methods are all tailored to their specific tasks, they all inspire our choice of a self-attending mechanism.



\section{Datasets} \label{Datasets}

\subsection{Corpora}
\label{sec:corpora}
We extract datasets from two corpora, namely the Penn Treebank (PTB) corpus \cite{marcus-etal-1993} and a subset (sections 000-760) of the third edition of the English Gigaword corpus \cite{gigaword}. For the PTB dataset, we use sections 22 and 23 for testing. For the Gigaword corpus, we use sections 700-760 for testing. For the remaining data, we randomly chose 10\% of them for development, and the other 90\% for training.

For each dataset, we consider a set of five target adverbs: \textit{too, again, also, still}, and \textit{yet}. We choose these five because they are commonly used adverbs that trigger presupposition. Since we are concerned with investigating the capacity of attentional deep neural networks in predicting the presuppositional effects in general, we frame the learning problem as a binary classification for predicting the presence of an adverbial presupposition (as opposed to the identity of the adverb).

On the Gigaword corpus, we consider each adverb separately, resulting in five binary classification tasks. This was not feasible for PTB because of its small size.

Finally, because of the commonalities between the adverbs in presupposing similar events, we create a dataset that unifies all instances of the five adverbs found in the Gigaword corpus, with a label ``1'' indicating the presence of any of these adverbs. 



\subsection{Data extraction process}
\label{balanced-dataset}
We define a sample in our dataset as a 3-tuple, consisting of a label (representing the target adverb, or `none' for a negative sample), a list of tokens we extract (before/after the adverb), and a list of corresponding POS tags \cite{Klein:2002:GCM:1073083.1073106}. In each sample, we also add a special token \textit{``@@@@''} right before the head word and the corresponding POS tag of the head word, both in positive and negative cases. We add such special tokens to identify the candidate context in the passage to the model. Figure~\ref{listing:example} shows a single positive sample in our dataset.

\begin{figure}  
\begin{lstlisting}[
frame=single,
basicstyle=\footnotesize,
breaklines=true,
language=DemoExample,
%label=listing:example
]
('still',
['The', 'Old', 'Granary', .../* 46 tokens omitted */...,'has', '@@@@', 'included', 'Bertrand', 'Russell', .../* 6 tokens omitted */... 'Morris'],
['DT', 'NNP', 'NNP', .../* 46 tokens omitted */..., 'VBZ', '@@@@', 'VBN', 'NNP', 'NNP', .../* 6 tokens omitted */... 'NNP'])
\end{lstlisting}
\caption{An example of an instance containing a presuppositional trigger from our dataset.}
\label{listing:example}
\end{figure}

We first extract positive contexts that contain a triggering adverb, then extract negative contexts that do not, controlling for a number of potential confounds. Our positive data consist of cases where the target adverb triggers presupposition by modifying a certain head word which, in most cases, is a verb. We define such head word as a \textit{governor} of the target adverb.

When extracting positive data, we scan through all the documents, searching for target adverbs. For each occurrence of a target adverb, we store the location and the governor of the adverb. Taking each occurrence of a governor as a pivot, we extract the 50 unlemmatized tokens preceding it, together with the tokens right after it up to the end of the sentence (where the adverb is)--with the adverb itself being removed. If there are less than 50 tokens before the adverb, we simply extract all of these tokens. In preliminary testing using a logistic regression classifier, we found that limiting the size to 50 tokens had higher accuracy than 25 or 100 tokens. As some head words themselves are stopwords, in the list of tokens, we do not remove any stopwords from the sample; otherwise, we would lose many important samples.

We filter out the governors of ``too" that have POS tags ``JJ'' and ``RB'' (adjectives and adverbs), because such cases corresponds to a different sense of ``too'' which indicates excess quantity and does not trigger presupposition (e.g., ``rely too heavily on'', ``it's too far from'').

After extracting the positive cases, we then use the governor information of positive cases to extract negative data. In particular, we extract sentences containing the same governors but not any of the target adverbs as negatives. In this way, models cannot rely on the identity of the governor alone to predict the class. This procedure also roughly balances the number of samples in the positive and negative classes.

For each governor in a positive sample, we locate a corresponding context in the corpus where the governor occurs without being modified by any of the target adverbs. We then extract the surrounding tokens in the same fashion as above. Moreover, we try to control position-related confounding factors by two randomization approaches: 1) randomize the order of documents to be scanned, and 2) within each document, start scanning from a random location in the document. Note that the number of negative cases might not be exactly equal to the number of negative cases in all datasets because some governors appearing in positive cases are rare words, and we're unable to find any (or only few) occurrences that match them for the negative cases.



\begin{table*}[ht!]
\centering
\begin{tabular*}{\textwidth}{@{\extracolsep{\fill}}l|rrr|rrr}
       & \multicolumn{3}{c|}{Training set} & \multicolumn{3}{c}{Test set} \\ \hline
Corpus & Positive       & Negative      & Total      & Positive     & Negative     & Total     \\ \hline
PTB       & 2,596       & 2,579        & 5,175 & 249 & 233 & 482\\
Gigaword yet     &  32,024      &     31,819    & 63,843   & 7950  & 7890 & 15840 \\
Gigaword too     &    55,827    &    29,918     & 85,745   & 13987 & 7514 & 21501 \\
Gigaword again   & 43,120       &    42,824     & 85,944   & 10935 & 10827 & 21762\\
Gigaword still   &   97,670     &    96,991     &  194,661 & 24509 & 24232 & 48741\\
Gigaword also    & 269,778      &    267,851    & 537,626  & 66878 & 66050 & 132928 \\
Gigaword all     &   498,415    &    491,173    & 989,588  & 124255 & 123078 & 247333\\   
\end{tabular*}
\caption{Number of training samples in each dataset. 
}
\label{tab:datasize}
\end{table*}

\section{Learning Model}
\label{sec:learn_model}
\begin{figure*}
    \centering
    \resizebox{\textwidth}{!}{   
        \includegraphics[scale=0.5]{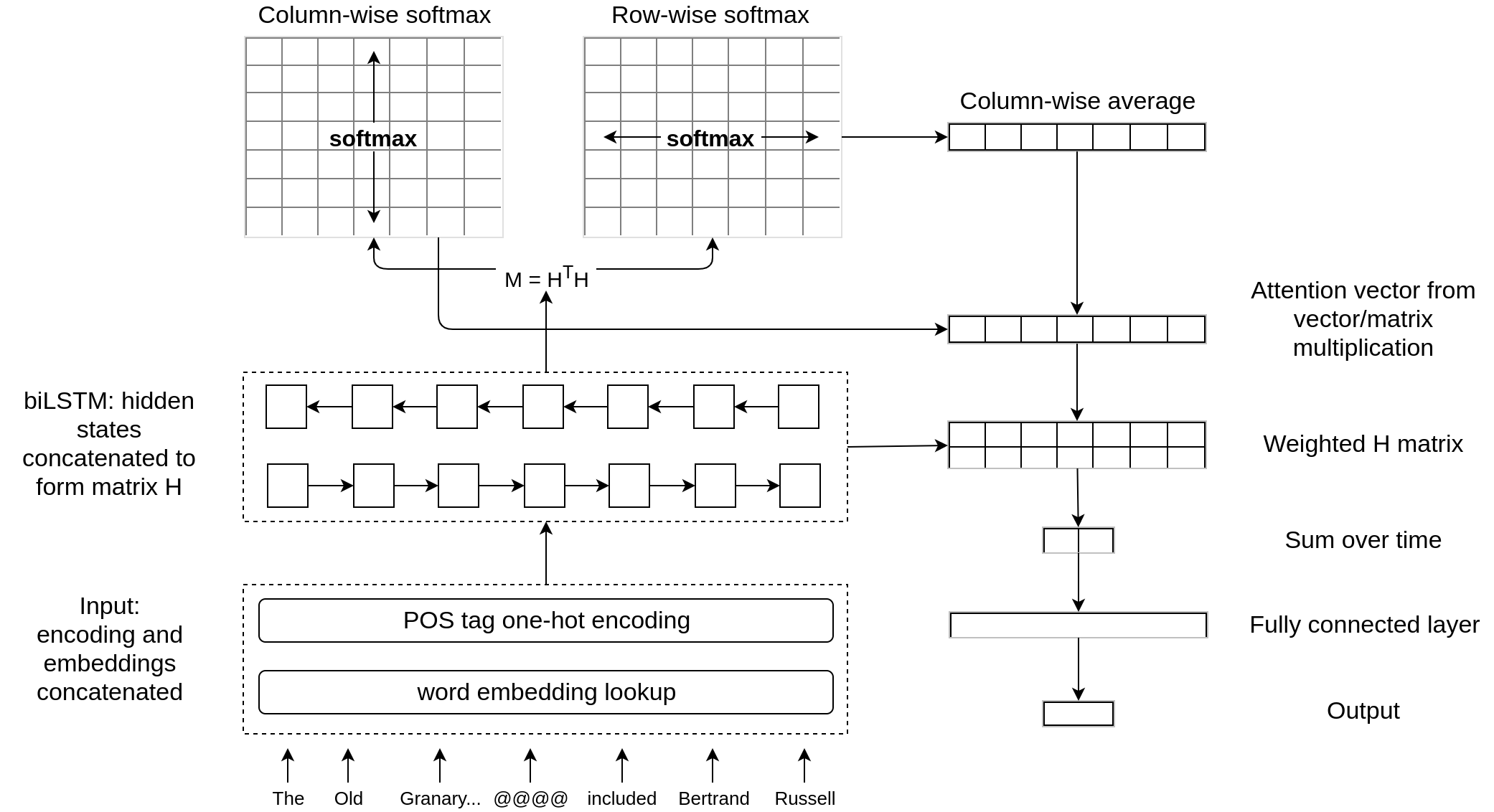}
    }
    \caption{Our weighted-pooling neural network architecture (WP). The tokenized input is embedded with pretrained word embeddings and possibly concatenated with one-hot encoded POS tags. The input is then encoded with a bi-directional LSTM, followed by our attention mechanism. The computed attention scores are then used as weights to average the encoded states, in turn connected to a fully connected layer to predict presupposition triggering.}
    \label{fig:model}
\end{figure*}

In this section, we introduce our attention-based model. At a high level, our model extends a bidirectional LSTM model by computing correlations between the hidden states at each timestep, then applying an attention mechanism over these correlations. Our proposed weighted-pooling (WP) neural network architecture is shown in Figure~\ref{fig:model}.

The input sequence $u = \{u_1, u_2, \dots, u_T\}$ consists of a sequence, of time length $T$, of one-hot encoded word tokens, where the original tokens are those such as in Listing~\ref{listing:example}. 
Each token $u_t$ is embedded with pretrained embedding matrix $W_e \in \mathbb{R}^{\vert V \vert \times d}$, where $\vert V \vert$ corresponds to the number of tokens in vocabulary $V$, and $d$ defines the size of the word embeddings. 
The embedded token vector $x_t \in \mathbb{R}^d$ is retrieved simply with $x_t = u_t W_e$. Optionally, $x_t$ may also include the token's POS tag. In such instances, the embedded token at time step $t$ is concatenated with the POS tag's one-hot encoding $p_t$: $x_t=u_t W_e || p_t$, where $||$ denotes the vector concatenation operator.

At each input time step $t$, a bi-directional LSTM~\cite{hochreiter1997long} encodes $x_t$ into hidden state $h_t \in \mathbb{R}^s$:
\beq
h_t = \left[ \overrightarrow{h_t} || \overleftarrow{h_t} \right]
\eeq
where $\overrightarrow{h_t} = f(x_t, h_{t-1})$ is computed by the forward LSTM, and $\overleftarrow{h_t} = f(x_t, h_{t+1})$ is computed by the backward LSTM. Concatenated vector $h_t$ is of size $2s$, where $s$ is a hyperparameter determining the size of the LSTM hidden states. Let matrix $H \in \mathbb{R}^{2s \times T}$ correspond to the concatenation of all hidden state vectors: 

\beq
H = [h_1 || h_2 || \dots || h_T].
\label{eq:h_encoded}
\eeq

Our model uses a form  of self-attention~\cite{paulus2017deep,vaswani2017attention}, where the input sequence acts as both the attention mechanism's query and key/value. Since the location of a presupposition trigger can greatly vary from one sample to another, and because dependencies can be long range or short range, we model all possible word-pair interactions within a sequence. We calculate the energy between all input tokens with a pair-wise matching matrix:
\beq
M = H^\top H
\eeq
where $M$ is a square matrix $\in \mathbb{R}^{T \times T}$. To get a single attention weight per time step, we adopt the attention-over-attention method~\cite{cui2017attention}. With matrix $M$, we first compute row-wise attention score $M_{ij}^r$ over $M$:
\beq
M_{ij}^r = \frac{exp(e_{ij})}{\sum_{t=1}^T exp(e_{it})}
\label{eq:row_attn}
\eeq
where $e_{ij} = M_{ij}$. $M^r$ can be interpreted as a word-level attention distribution over all other words. Since we would like a single weight per word, we need an additional step to aggregate these attention scores. Instead of simply averaging the scores, we follow \cite{cui2017attention}'s approach which learns the aggregation by an additional attention mechanism. We compute column-wise softmax $M_{ij}^c$ over $M$:
\beq
M_{ij}^c = \frac{exp(e_{ij})}{\sum_{t=1}^T exp(e_{tj})}
\eeq

The columns of $M^r$ are then averaged, forming vector $\beta \in \mathbb{R}^T$. Finally, $\beta$ is multiplied with the column-wise softmax matrix $M^c$ to get attention vector $\alpha$:

\beq
\alpha = M^{r\top} \beta.
\label{eq:attn_vector}
\eeq

Note Equations~(\ref{eq:h_encoded}) to (\ref{eq:attn_vector}) have described how we derived an attention score over our input without the introduction of any new parameters, potentially minimizing the computational effect of our attention mechanism. 

As a last layer to their neural network, Cui et al.~\shortcite{cui2017attention} sum over $\alpha$ to extract the most relevant input. However, we use $\alpha$ as weights to combine all of our hidden states $h_t$:
\beq
c = \sum_{t=1}^T \alpha_t h_t
\eeq
where $c \in \mathbb{R}^s$. We follow the pooling with a dense layer $z = \sigma(W_z c + b_z)$, where $\sigma$ is a non-linear function, matrix $W_z \in \mathbb{R}^{64 \times s}$ and vector $b_z \in \mathbb{R}^{64}$ are learned parameters. 
The presupposition trigger probability is computed with an affine transform followed by a softmax:
\beq
\hat{y} = \text{softmax}(W_o z + b_o)
\eeq
where matrix $W_o \in \mathbb{R}^{2 \times 64}$ and vector $b_o \in \mathbb{R}^2$ are learned parameters. The training objective minimizes:
\beq
J(\theta) = \frac{1}{m} \sum_{t=1}^m E(\hat{y}, y)
\eeq
where $E(\cdot~,\cdot)$ is the standard cross-entropy.

\begin{table*}[ht]
\centering
\begin{tabular*}{\textwidth}{@{\extracolsep{\fill}}lcccccccc}
\cmidrule(l){3-9}
                        & \multicolumn{1}{l}{}          & \multicolumn{7}{c}{Accuracy}                                                                                                              \\ \cmidrule(l){3-9} 
                        & \multicolumn{1}{l}{}          & \multicolumn{1}{c|}{WSJ} & \multicolumn{6}{c}{Gigaword}                                                                       \\ \midrule
Models                  & \multicolumn{1}{l|}{Variants} & \multicolumn{1}{c|}{All adverbs}    & All adverbs    & Also           & Still  & Again        & Too            & Yet                       \\ \midrule
MFC                     & \multicolumn{1}{c|}{-}        & \multicolumn{1}{c|}{51.66}          & 50.24          & 50.32          & 50.29  & 50.25        & 65.06          & 50.19                    \\ \midrule
\multirow{2}{*}{LogReg} & \multicolumn{1}{c|}{+ POS}    & \multicolumn{1}{c|}{52.81}          & 53.65          & 52.00           & 56.36 & 59.49         & 69.77          & \textbf{61.05}            \\
                        & \multicolumn{1}{c|}{- POS}    & \multicolumn{1}{c|}{54.47}          & 52.86          & 56.07           & 55.29 & 58.60         & 67.60          & 58.60                    \\ \midrule
\multirow{2}{*}{CNN}    & \multicolumn{1}{c|}{+ POS}    & \multicolumn{1}{c|}{58.84}          & 59.12          & 61.53           & 59.54 & 60.26          & 67.53          & 59.69                    \\
                        & \multicolumn{1}{c|}{- POS}    & \multicolumn{1}{c|}{62.16}          & 57.21          & 59.76           & 56.95 & 57.28         & 67.84          & 56.53                    \\ \midrule
\multirow{2}{*}{LSTM}   & \multicolumn{1}{c|}{+ POS}    & \multicolumn{1}{c|}{74.23}          & 60.58          & 81.48  & 60.72 & \textbf{61.81}          & \textbf{69.70} & 59.13                    \\
                        & \multicolumn{1}{c|}{- POS}    & \multicolumn{1}{c|}{73.18}          & \textbf{58.86}          & 81.16  & \textbf{58.97} & \textbf{59.93}         & \textbf{68.32}          & 55.71                    \\ \midrule
\multirow{2}{*}{WP}     & \multicolumn{1}{c|}{+ POS}    & \multicolumn{1}{c|}{\textbf{76.09}} & \textbf{60.62} & \textbf{82.42}           & \textbf{61.00} &  61.59 & 69.38          & 57.68           \\
                        & \multicolumn{1}{c|}{- POS}    & \multicolumn{1}{c|}{\textbf{74.84}} & \textbf{58.87} & \textbf{81.64}           & \textbf{59.03} & 58.49 & \textbf{68.37} & \textbf{56.68}  \\ \bottomrule
\end{tabular*}

\caption{Performance of various models, including our weighted-pooled LSTM (WP). MFC refers to the  most-frequent-class baseline, LogReg is the logistic regression baseline. LSTM and CNN correspond to strong neural network baselines. Note that we bold the performance numbers for the best performing model for each of the ``+ POS'' case and the ``- POS'' case.}
\label{tab:results}
\end{table*}

\section{Experiments}
We compare the performance of our WP model against several models which we describe in this section. We carry out the experiments on both datasets described in Section~\ref{Datasets}. We also investigate the impact of POS tags and attention mechanism on the models' prediction accuracy.

\subsection{Baselines}
\label{sec:baselines}
We compare our learning model against the following systems. The first is the most-frequent-class baseline \textbf{(MFC)} which simply labels all samples with the most frequent class of 1. The second is a logistic regression classifier \textbf{(LogReg)}, in which the probabilities describing the possible outcomes of a single input $x$ is modeled using a logistic function. We implement this baseline classifier with the scikit-learn package \cite{scikit-learn}, with a CountVectorizer including bi-gram features. All of the other hyperparameters are set to default weights.  

The third is a variant LSTM recurrent neural network as introduced in \cite{Graves13}. The input is encoded by a bidirectional LSTM like the WP model detailed in Section~\ref{sec:learn_model}. Instead of a self-attention mechanism, we simply mean-pool matrix $H$, the concatenation of all LSTM hidden states, across all time steps. This is followed by a fully connected layer and softmax function for the binary classification. Our WP model uses the same bidirectional LSTM as this baseline LSTM, and has the same number of parameters, allowing for a fair comparison of the two models. Such a standard LSTM model represents a state-of-the-art language model, as it outperforms more recent models on language modeling tasks when the number of model parameters is controlled for~\cite{melis2017evaluation}.

For the last model, we use a slight variant of the CNN sentence classification model of \cite{Kim2014} based on the Britz tensorflow implementation\footnote{http://www.wildml.com/2015/12/implementing-a-cnn-for-text-classification-in-tensorflow/}. 



\subsection{Hyperparameters \& Additional Features}
After tuning, we found the following hyperparameters to work best: 64 units in fully connected layers and 40 units for POS embeddings. We used dropout with probability 0.5 and mini-batch size of 64.
 
For all models, we initialize word embeddings with word2vec~\cite{mikolov2013distributed} pretrained embeddings of size 300. Unknown words are randomly initialized to the same size as the word2vec embeddings. In early tests on the development datasets, we found that our neural networks would consistently perform better when fixing the word embeddings. All neural network performance reported in this paper use fixed embeddings. 

Fully connected layers in the LSTM, CNN and WP model are regularized with dropout~\cite{srivastava2014dropout}.
The model parameters for these neural networks are fine-tuned with the Adam algorithm~\cite{kingma2015method}.
To stabilize the RNN training gradients~\cite{pascanu2013difficulty}, we perform gradient clipping for gradients below threshold value -1, or above 1. To reduce overfitting, we stop training if the development set does not improve in accuracy for 10 epochs. All performance on the test set is reported using the best trained model as measured on the development set.

In addition, we use the CoreNLP Part-of-Speech (POS) tagger \cite{manning-EtAl:2014:P14-5} to get corresponding POS features for extracted tokens. In all of our models, we limit the maximum length of samples and POS tags to 60 tokens. For the CNN, sequences shorter than 60 tokens are zero-padded.

\section{Results}
Table~\ref{tab:results} shows the performance obtained by the different models with and without POS tags. Overall, our attention model WP outperforms all other models in 10 out of 14 scenarios (combinations of datasets and whether or not POS tags are used). Importantly, our model outperforms the regular LSTM model without introducing additional parameters to the model, which highlights the advantage of WP's attention-based pooling method. For all models listed in Table~\ref{tab:results}, we find that including POS tags benefits the detection of adverbial presupposition triggers in Gigaword and PTB datasets. Note that, in Table~\ref{tab:results}, we bolded accuracy figures that were within 0.1\% of the best performing WP model as McNemar's test did not show that WP significantly outperformed the other model in these cases (\textit{p} value > 0.05).

Table~\ref{tab:confusion} shows the confusion matrix for the best performing model (WP,+POS). The small differences in the off-diagonal entries inform us that the model misclassifications are not particularly skewed towards the presence or absence of presupposition triggers.

\begin{table}[ht]
\centering
\begin{tabular}{ll|rr}
                                 &          & \multicolumn{2}{c}{\textbf{Predicted}} \\
\parbox[t]{2mm}{\multirow{2}{*}{\rotatebox[origin=c]{90}{\textbf{Actual}}}}                                 &          & Absence           & Presence           \\\cline{2-4}
 & Absence  & 54,658              & 11,961               \\
 & Presence & 11,776               & 55,006               \\
\end{tabular}
\caption{Confusion matrix for the best performing model, predicting the presence of a presupposition trigger or the absence of such as trigger.}
\label{tab:confusion}
\end{table}

\begin{table}[ht!]
\centering
\begin{tabular}{l|rr}
			   & \textbf{WP Cor.} & \textbf{WP Inc.} \\\hline
\textbf{LSTM Cor.}   & 101,443     & 6,819 \\
\textbf{LSTM Inc.} & 8,016       & 17,123\\
\end{tabular}
\caption{Contingency table for correct (cor.) and incorrect (inc.) predictions between the LSTM baseline and the attention model (WP) on the Giga\_also dataset.}
\label{tab:agreement}
\end{table}

The contingency table, shown in Table~\ref{tab:agreement}, shows the distribution of agreed and disagreed classification. 

\section{Analysis}
Consider the following pair of samples that we randomly choose from the PTB dataset (shortened for readability):

\begin{enumerate}
	\item \label{exe4} ...Taped just as the market closed yesterday , it offers Ms. Farrell advising , " We view the market here as going through a relatively normal cycle ... . We \textbf{continue} to feel that the stock market is the @@@@ place to be for long-term appreciation
	
	\item \label{exe5} ...More people are remaining independent longer presumably because they are better off physically and financially . Careers count most for the well-to-do many affluent people @@@@ place personal success and money above family
\end{enumerate}

In both cases, the head word is \textit{place}. 
In Example \ref{exe4}, the word \textit{continue} (emphasized in the above text) suggests that adverb \textit{still} could be used to modify head word \textit{place} (i.e., \textit{... the stock market is still the place ...}). Further, it is also easy to see that \textit{place} refers to \textit{stock market}, which has occurred in the previous context. Our model correctly predicts this sample as containing a presupposition, this despite the complexity of the coreference across the text. 

In the second case of the usage of the same main head word \textit{place} in Example~\ref{exe5}, our  model falsely predicts the presence of a presupposition. However, even a human could read the sentence as ``many people still place personal success and money above family''. This underlies the subtlety and difficulty of the task at hand. The long-range dependencies and interactions within sentences seen in these examples are what motivate the use of the various deep non-linear models presented in this work, which are useful in detecting these coreferences, particularly in the case of attention mechanisms.



\section{Conclusion} 

In this work, we have investigated the task of predicting adverbial presupposition triggers and introduced several datasets for the task. Additionally, we have presented a novel weighted-pooling attention mechanism which is incorporated into a recurrent neural network model for predicting the presence of an adverbial presuppositional trigger. Our results show that the model outperforms the CNN and LSTM, and does not add any additional parameters over the standard LSTM model. This shows its promise in classification tasks involving capturing and combining relevant information from multiple points in the previous context. 

In future work, we would like to focus more on designing models that can deal with and be optimized for scenarios with severe data imbalance
. We would like to also explore various applications of presupposition trigger prediction in language generation applications, as well as additional attention-based neural network architectures.

\section*{Acknowledgements}
The authors would like to thank the reviewers for their valuable comments. This work was supported by the Centre de Recherche d'Informatique de Montréal (CRIM), the Fonds de Recherche du Québec – Nature et Technologies (FRQNT) and the Natural Sciences and Engineering Research Council of Canada (NSERC).

\bibliographystyle{acl_natbib}
\bibliography{reference}
\end{document}